\documentclass[lettersize,journal]{IEEEtran}
\IEEEoverridecommandlockouts

\usepackage{amsmath}
\usepackage{graphicx}
\usepackage{booktabs}
\usepackage{soul}
\usepackage{xcolor}
\usepackage{multirow}
\usepackage{enumitem}
\usepackage{stfloats}

\definecolor{CJ_red}{HTML}{d95e5f}
\definecolor{CJ_green}{HTML}{3e8a2f}
\definecolor{CJ_blue}{HTML}{4c96d9}
\definecolor{CJ_yellow}{HTML}{c69527}
\definecolor{CJ_purple}{HTML}{9e4588}

\makeatletter
\renewcommand\footnoterule{%
  \kern -3pt
  \hrule width \columnwidth
  \kern 2.6pt
}
\makeatother

\begin{document}

\title{From Inference Efficiency to Embodied Efficiency:\\Revisiting Efficiency Metrics for Vision-Language-Action Models}
\author{
\IEEEauthorblockN{
Zhuofan Li\textsuperscript{1*},
Hongkun Yang\textsuperscript{1*},
Zhenyang Chen\textsuperscript{2*},
Yangxuan Chen\textsuperscript{1},
Yingyan (Celine) Lin\textsuperscript{2+},
Chaojian Li\textsuperscript{1+}
}
\IEEEauthorblockA{
\textsuperscript{1}\textit{Hong Kong University of Science and Technology}\\
\textsuperscript{2}\textit{Georgia Institute of Technology}
}\thanks{\textsuperscript{*}These authors contributed equally.}
\thanks{\textsuperscript{+}Correspondence to: Yingyan (Celine) Lin (celine.lin@gatech.edu) and Chaojian Li (chaojian@ust.hk).}
}

\maketitle
\begin{abstract}

Vision-Language-Action (VLA) models have recently enabled embodied agents to perform increasingly complex tasks by jointly reasoning over visual, linguistic, and motor modalities. However, we find that the prevailing notion of ``efficiency'' in current VLA research, characterized by parameters, FLOPs, or token decoding throughput, does not reflect actual performance on robotic platforms. In real-world execution, efficiency is determined by system-level embodied behaviors such as task completion time, trajectory smoothness, cumulative joint rotation, and motion energy. Through controlled studies across model compression, token sparsification, and action sequence compression, we make several observations that challenge common assumptions. (1) Methods that reduce computation under conventional metrics often increase end-to-end execution cost or degrade motion quality, despite maintaining task success rates.
(2) System-level embodied efficiency metrics reveal performance differences in the learned action policies that remain hidden under conventional evaluations.
(3) Common adaptation methods such as in-context prompting or supervised fine-tuning show only mild and metric-specific improvements in embodied efficiency. While these methods can reduce targeted embodied-efficiency metrics such as jerk or action rate, the resulting gains may come with trade-offs in other metrics, such as longer completion time. Taken together, our results suggest that conventional inference efficiency metrics can overlook important aspects of embodied execution. Incorporating embodied efficiency provides a more complete view of policy behavior and practical performance, enabling fairer and more comprehensive comparisons of VLA models.

\end{abstract}

\section{Introduction}
\label{sec:intro}

\begin{figure}[t] 
    \centering \includegraphics[width=1\linewidth]{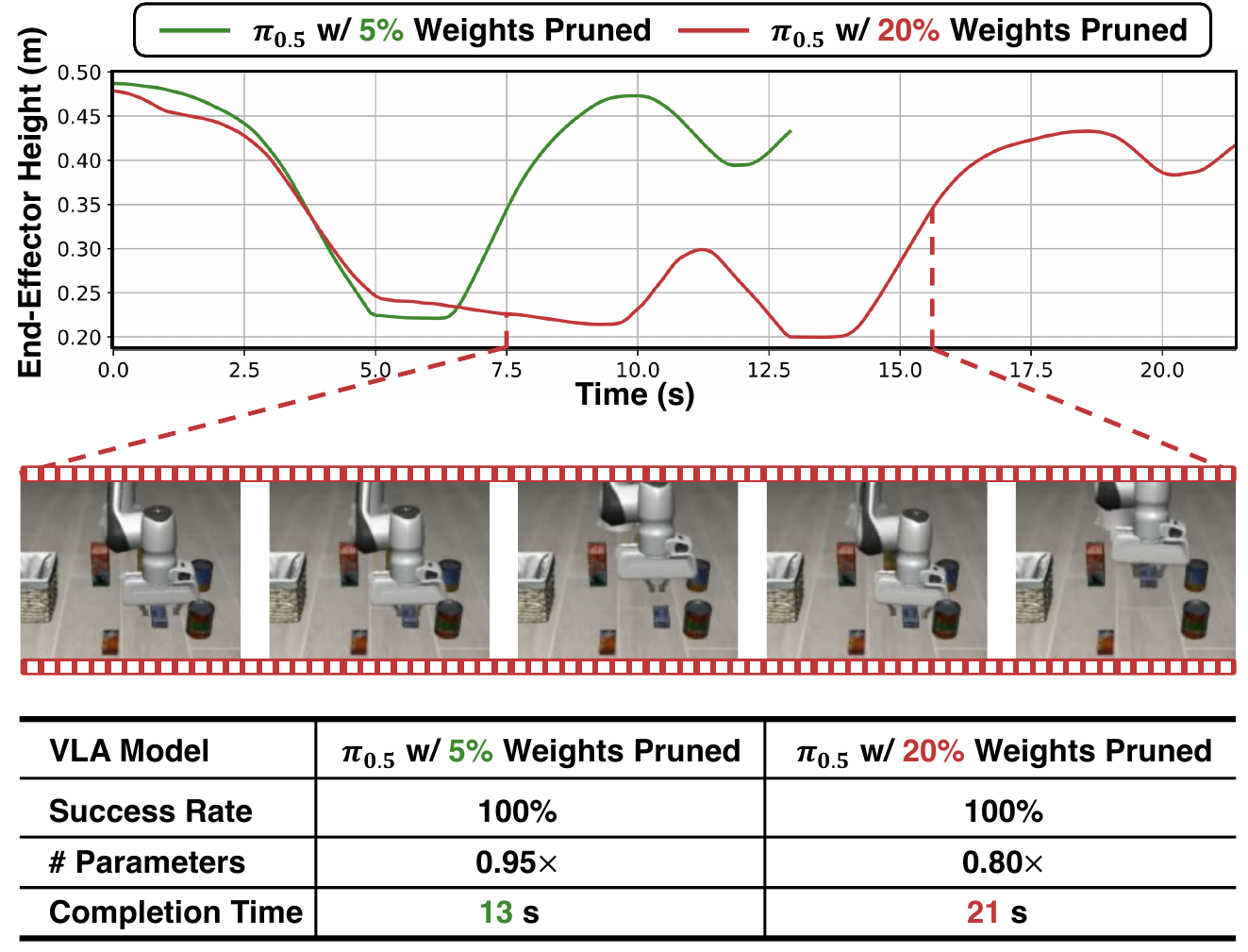} 
        \vspace{-1.8em}
\caption{Comparison of $\pi_{0.5}$~\cite{black2025pi05} VLA models with 5\% and 20\% of weights pruned using magnitude pruning~\cite{han2015learning}, evaluated on a Libero~\cite{liu2023libero} task with the instruction \textit{``pick up the cream cheese and place it in the basket.''} The comparison includes task success rate (averaged over 50 test rollout trajectories, following~\cite{liu2023libero}), parameter count, completion time, and end-effector height over time. Notably, in this extreme case, although the 20\% pruned model achieves a larger parameter reduction without any drop in success rate, it fails to execute the grasp smoothly, as shown in the zoomed-in sequence. This results in a longer completion time and therefore higher overall system energy consumption. This makes the model \textbf{``inference-efficient''} but not \textbf{``embodied-efficient''}.}
\vspace{-0.5em}
    \label{fig:teaser} 
\end{figure}

\begin{figure*}[t] 
    \centering \includegraphics[width=1\linewidth]{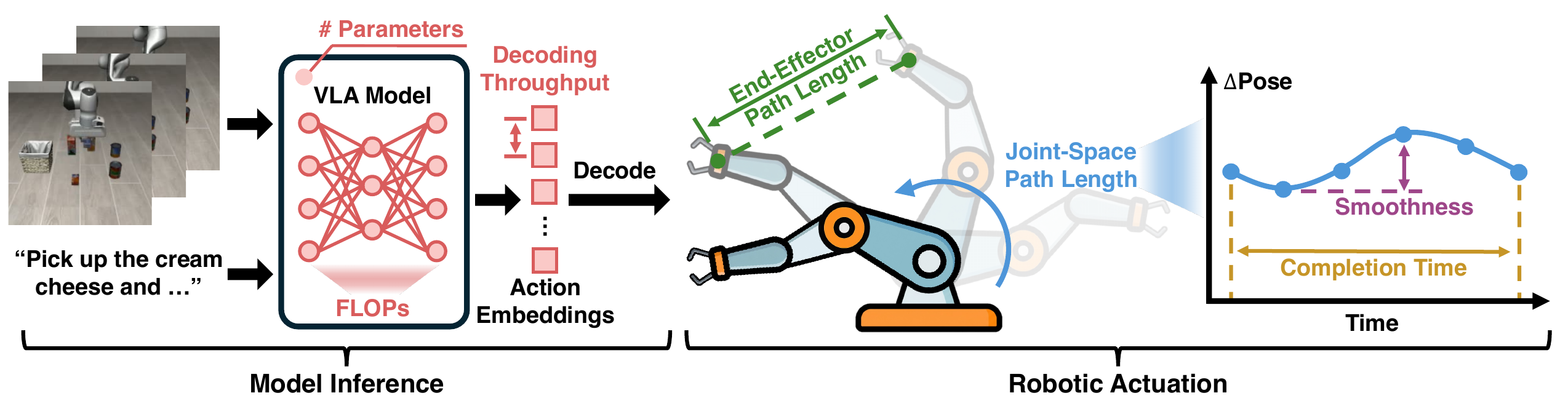} 
        \vspace{-2em}
\caption{An overview of the deployment of a VLA model on embodied robotic platforms. Conventional model \textbf{inference efficiency} metrics (e.g., \textcolor{CJ_red}{\textbf{number of parameters}}, \textcolor{CJ_red}{\textbf{FLOPs}}, and \textcolor{CJ_red}{\textbf{decoding throughput}}) apply only to the model inference stage, where the VLA model processes captured images and language instructions to output corresponding action tokens. However, assessing \textbf{embodied efficiency} for the robotic actuation stage requires a different set of metrics, including \textcolor{CJ_green}{\textbf{end-effector path length}}, \textcolor{CJ_blue}{\textbf{joint-space path length}}, \textcolor{CJ_purple}{\textbf{action smoothness}}, and \textcolor{CJ_yellow}{\textbf{task completion time}}.}
    \label{fig:overview} 
\end{figure*}

Vision-Language-Action (VLA) models have recently emerged as a promising paradigm for enabling embodied agents to perform complex, instruction-conditioned behaviors by jointly reasoning over visual observations, natural language goals, and robot states
~\cite{pmlr-v229-zitkovich23a,kim_openvla_2024,black2024pi0,black2025pi05}. 
As VLA models continue to increase in scale and capability, their ability to execute behaviors efficiently becomes increasingly important for practical deployment, especially in latency- or energy-constrained settings~\cite{guan2025efficientvisionlanguageactionmodelsembodied,fang2025sqap,ma2025running,jiang2025better}. However, unlike text- or image-only tasks, the performance of embodied agents is determined not only by the efficiency of model inference, but also by the physical behavior of the robotic platform induced by the model’s output actions. Specifically, the power consumption associated with \ul{model inference} and the power consumption associated with \ul{robotic actuation} are often of the same order of magnitude. For example, when running the LeRobot example on an NVIDIA Jetson platform~\cite{lerobot_nvidia_2024}, the total power draw of the servos in the Koch v1.1 kit~\cite{koch_v1_1_follower_robotis_2025,koch_v1_1_leader_robotis_2025} under their rated torque is approximately 22~W, while the maximum power budget of the corresponding Jetson Orin NX edge GPU~\cite{jetson_orin_nx_datasheet_nvidia_2022} is 20~W. A similar trend holds for more complex embodied systems: a Franka Emika robot arm and its controller~\cite{franka_datasheet_2020} can consume up to 430~W, whereas an NVIDIA H100 NVL GPU~\cite{nvidia_h100_datasheet_2022} has a maximum power consumption of around 400~W. These examples highlight that \textbf{improving inference efficiency alone does not guarantee better system-level embodied efficiency}, as actuation costs can dominate or offset model-level gains.

Despite the importance of the aforementioned robotic actuation in determining overall VLA system efficiency, existing work on efficient VLA does not explicitly incorporate actuation cost when designing compression techniques. Instead, the prevailing approach is to apply efficient AI methods originally developed for large language or vision models, such as pruning~\cite{li_sp-vla_2025, fang_sqap-vla_2025}, quantization~\cite{fang_sqap-vla_2025}, token sparsification~\cite{jiang_better_2025}, or lightweight inference architectures~\cite{wen2025tinyvlafastdataefficientvisionlanguageaction, liu2024robomambaefficientvisionlanguageactionmodel, shukor2025smolvlavisionlanguageactionmodelaffordable}, and to evaluate their effectiveness using conventional AI efficiency metrics such as parameter count~\cite{wen2025tinyvlafastdataefficientvisionlanguageaction, shukor2025smolvlavisionlanguageactionmodelaffordable}, floating point operations (FLOPs)~\cite{yue2024deervladynamicinferencemultimodal}, or decoding throughput (e.g., tokens per second)~\cite{kim2025finetuningvisionlanguageactionmodelsoptimizing}. However, these metrics capture only the efficiency of \ul{model inference} and fail to reflect the system-level \ul{robotic actuation} behavior that ultimately determines embodied performance. For example, as shown in Fig.~\ref{fig:teaser}, two models with identical task success rates (i.e., 100\%) can exhibit drastically different task completion times (13 s vs. 21 s). In this case, the model with 20\% of its weights pruned incurs higher overall system energy (i.e., less \textbf{``embodied-efficient''}) expenditure than the 5\%-pruned model, despite having fewer parameters and no increase in FLOPs or decoding throughput (i.e., more \textbf{``inference-efficient''}).

To comprehensively study the mismatch between the prevailing notion of ``inference efficiency'' and the desired ``embodied efficiency'', we examine several commonly used efficiency-improving techniques. These techniques are evaluated using multiple embodied-efficiency metrics, as depicted in Fig.~\ref{fig:overview}. We conduct this analysis across three domains of VLA design:
(1) \textbf{model} compression (e.g., weight pruning and quantization)~\cite{li_sp-vla_2025, fang_sqap-vla_2025}, (2) \textbf{token} sparsification (e.g., visual token pruning)~\cite{yang_efficientvla_2025, li_sp-vla_2025}, and (3) \textbf{action} sequence compression (e.g., reducing temporal redundancy in control signals)~\cite{pertsch2025fastefficientactiontokenization}. Across these domains, our analysis reveals several insights that challenge prevailing assumptions:

\begin{itemize}[leftmargin=*]
    \item \textbf{Reducing computational inference cost does not necessarily improve embodied execution efficiency.} 
    Techniques that decrease FLOPs or tokens per forward pass can increase end-to-end task duration or degrade motion smoothness, even when task success rates are preserved. In particular, pruning 5\% of the weights of a commonly used $\pi_{0}$ VLA model~\cite{black2024pi0} results in a 13.6\% longer task completion time and a 46.2\% increase in end-effector path length on the Bridge benchmark~\cite{walke2024bridgedatav2datasetrobot}.

    \item \textbf{System-level embodied metrics reveal differences that conventional evaluations overlook.} Policies learned by different VLA models that appear equivalent under accuracy- or throughput-based evaluations may differ considerably in the motion dynamics and energy usage they induce. Specifically, $\pi_{0}$~\cite{black2024pi0} and $\pi_{0}$-FAST~\cite{pertsch2025fastefficientactiontokenization} achieve identical success rates on Libero-Object~\cite{liu2023libero}, but $\pi_{0}$-FAST exhibits a 1.5\% shorter task completion time along with 34.5\% higher jerkiness, implying a more aggressive policy learned by $\pi_{0}$-FAST.
    
    \item \textbf{Common adaptation methods show mild and metric-specific improvements in embodied efficiency.} We find that directly applying supervised fine-tuning or in-context prompting can reduce targeted embodied-efficiency metrics such as jerk or action rate, but these gains are often partial and may come with trade-offs in other metrics. For example, adding an auxiliary jerk-based loss during supervised fine-tuning of the $\pi_{0.5}$~\cite{black2025pi05} model on Libero-Object~\cite{liu2023libero} reduces jerkiness by 13.2\% without changing end-effector path length, while in-context prompting can produce larger jerk reductions of up to 25.8\% on Libero-Goal, although sometimes at the cost of longer completion time.
\end{itemize}

\section{Related Work}
\label{sec:related_works}

\subsection{VLA for Robotic Control}
\label{sec:related_works:VLA}

VLA models extend the multimodal capabilities of Vision-Language Models (VLMs) to robotic actuation and have rapidly become one of the dominant paradigms for generalist robotic control tasks. In particular, the Robotics Transformer (RT) family~\cite{brohan2022rt1,pmlr-v229-zitkovich23a,oneill2023openx,belkhale2024rth} demonstrates that large transformer policies trained on diverse, language-conditioned robot demonstrations can unify hundreds of manipulation tasks. 
To improve temporal coherence of action trajectories and execution efficiency, $\pi_{0}$~\cite{black2024pi0} and $\pi_{0.5}$~\cite{black2025pi05} replace the autoregressive head design with generative action decoders based on diffusion and flow matching. Nevertheless, due to their large transformer backbones, current VLA models entail substantial computation and data movement during inference, leading to high latency and memory usage~\cite{kawaharazuka2025vision}. For practical deployment on resource-constrained robotic platforms that require rapid decision-making under limited power budgets, improving the efficiency of VLAs remains a key challenge.

\subsection{Efficient VLA}
\label{sec:related_works:efficient_vla}

In the \textbf{model} domain, 
OpenVLA~\cite{kim_openvla_2024} reduces the memory footprint during inference by quantizing weights from \texttt{bfloat16} to \texttt{int8} and \texttt{int4}. 
QAIL~\cite{park2024quantizationawareimitationlearningresourceefficientrobotic} proposes a quantization-aware fine-tuning scheme to jointly reduce memory consumption and latency per inference step. Beyond quantization, GLUESTICK~\cite{jabbour2025don} explores recovering task success rate after structured pruning of less important weights in VLA models. In the \textbf{token} domain, prior works have investigated pruning less informative tokens to reduce prefilling cost. 
SP-VLA~\cite{li_sp-vla_2025} and SQAP-VLA~\cite{fang_sqap-vla_2025} prune visual tokens to accelerate the perception-to-action pipeline. 
Several studies also design new VLA pipelines that effectively compress both the model and token representations. For instance, 
RoboMamba~\cite{liu2024robomambaefficientvisionlanguageactionmodel} replaces the transformer backbone with a state-space model to facilitate faster token decoding. Beyond compression in the model and token domains, several efforts target the \textbf{action} representation domain to more compactly represent action sequences. 
FAST~\cite{pertsch2025fastefficientactiontokenization} leverages a discrete cosine transform to reduce temporal redundancy in action chunks.
RynnVLA-001~\cite{jiang2025rynnvla} adopts a variational autoencoder for action compression, reducing the dimensionality of the action space. 
Despite these advances toward more efficient VLA models, most existing work overlooks the overall energy efficiency of the full closed-loop system, encompassing both model inference and the resulting robotic actuation. 

\begin{figure*}[!b] 
    \centering \includegraphics[width=1\linewidth]{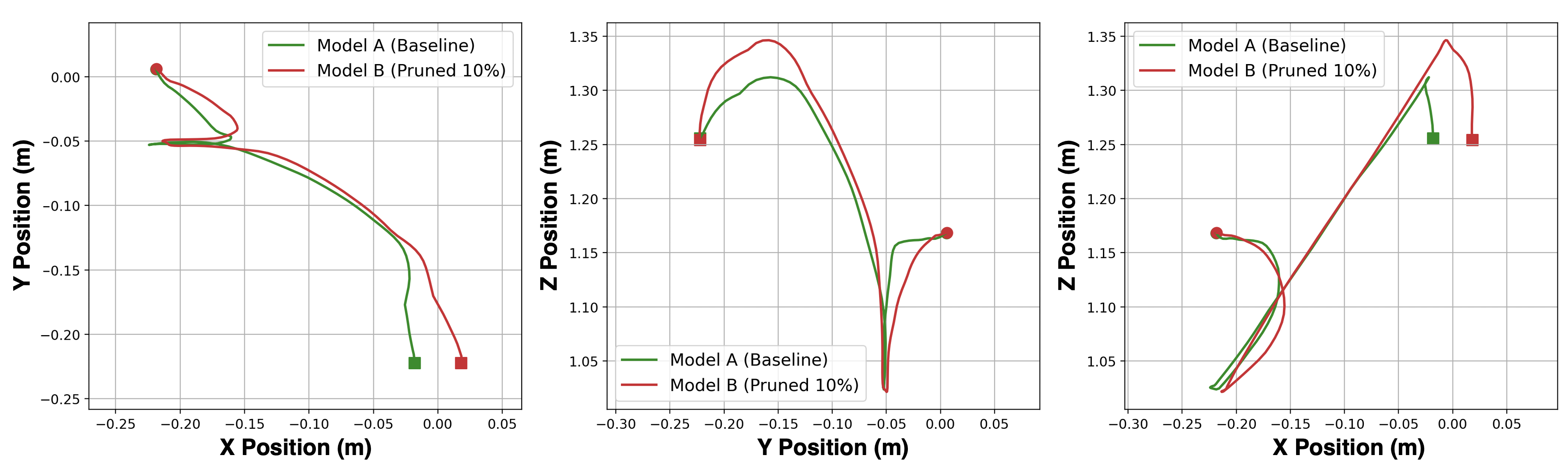} 
        \vspace{-2em}
\caption{Visualization of the trajectories of Model A (a baseline $\pi_{0}$) and Model B (a $\pi_{0}$ model with 10\% of weights pruned) when performing the 4th task of the Libero-Goal suite~\cite{liu2023libero}. The trajectories are projected onto the X--Y, Y--Z, and X--Z planes for improved visualization clarity, revealing that Model B induces a longer path.}
    \label{fig:example_traj} 
\end{figure*}

\section{Preliminaries}
\label{sec:preliminaries}

\subsection{From Model Inference to Robotic Actuation}
\label{sec:preliminaries:pipeline}

Following the formulation in~\cite{brohan2022rt1}, for the deployment of VLA models on robotic platforms, we model the closed-loop execution of a language-conditioned robot as a sequential decision-making process $(\mathcal{S}, \mathcal{O}, \mathcal{A}, \mathcal{P})$, where $\mathcal{S}$, $\mathcal{O}$, $\mathcal{A}$, and $\mathcal{P}$ denote the state, observation, action spaces, and transition function, respectively. At timestep $t$, the environment is in state $s_t \in \mathcal{S}$, the robot receives an observation $o_t \in \mathcal{O}$ (e.g., RGB images, as depicted on the left side of Fig.~\ref{fig:overview}), and the VLA model $\pi_\theta$ outputs an action $a_t \in \mathcal{A}$ conditioned on the observation $o_t$ and a language goal $g$, i.e.,
\begin{equation}
a_t \sim \pi_\theta(a_t \mid o_t, g).
\end{equation}
The environment then evolves according to the transition kernel $s_{t+1} \sim \mathcal{P}(\cdot \mid s_t, a_t)$, and low-level controllers map the high-level actions $a_t$ to joint torques and resulting motions. This formulation provides the basis for analyzing the embodied efficiency of VLA-driven robotic systems.

\subsection{Evaluation Metrics for Embodied Efficiency}
\label{sec:preliminaries:metrics}

To evaluate the embodied efficiency of the inference-to-actuation process, we adopt metrics that capture both model inference behavior and robotic actuation performance.
Specifically, the adopted embodied efficiency metrics fall into the following two categories.
(1) \textbf{Task duration metrics}, which quantify the accumulated execution cost during embodied actuation driven by VLA inference:
\begin{itemize}[leftmargin=*]
    \item \textbf{Task completion time} $\tau$ is defined as $\tau = T/f$,
    where $T$ is the number of timesteps and $f$ is the control frequency. A smaller $\tau$ indicates higher execution efficiency, as it measures the total time during which energy is consumed for model inference and robotic actuation.

    \item \textbf{End-effector path length} $L_{\text{ee}}$ is defined as $L_{\text{ee}} = \sum_{t=1}^{T-1} \bigl\|p_{t+1} - p_t\bigr\|_2$,
    where $p_{t}$ denotes the end-effector pose (e.g., the robot's gripper in Fig.~\ref{fig:overview}) at timestep $t$. A shorter path $L_{\text{ee}}$ indicates less actuator travel and thus higher efficiency.

    \item \textbf{Joint-space path length} $L_{\text{joint}}$ is defined as $L_{\text{joint}} = \sum_{t=1}^{T-1} \bigl\|q_{t+1} - q_t\bigr\|_2$,
    where $q_{t}$ represents the joint configuration at timestep $t$ (e.g., the blue arrow in Fig.~\ref{fig:overview}). A smaller $L_{\text{joint}}$ corresponds to reduced mechanical work and therefore higher efficiency.
\end{itemize}
(2) \textbf{Motion smoothness metrics}, which measure jerk or jitter that arises during robotic actuation due to unstable VLA action outputs or physical constraints:
\begin{itemize}[leftmargin=*]
    \item \textbf{Average jerk L2 norm} $J$ is defined as $J = \frac{f^4}{T-2} \sum_{t=2}^{T-1} \|\dot{q}_{t+1} - 2\dot{q}_t + \dot{q}_{t-1}\|_2^2,$
    which averages the L2 norm of the discrete second-order difference of joint configurations. A higher $J$ reflects more frequent accelerations and decelerations, which generally incur higher instantaneous power and dissipative losses~\cite{arachchige2025sailfasterthandemonstrationexecutionimitation}.

    \item \textbf{Average action rate} $R$ is defined as $R = \frac{1}{T-1} \sum_{t=1}^{T-1} \bigl\|a_{t+1} - a_t\bigr\|_2$,
    which measures how rapidly high-level actions change across timesteps. A higher $R$ indicates more aggressive command variations and thus typically lower energy efficiency.
\end{itemize}

\noindent To reduce random variation, we evaluate each policy across $N$ episodes and compute the overall task success rate $\text{SR}$ as $\text{SR} = \frac{1}{N} \sum_{i=1}^N \mathbf{1}[\text{succ}_i]$,
where $\mathbf{1}[\text{succ}_i]$ is the success indicator for the $i$-th episode. Similarly, all the aforementioned metrics are averaged over the successful episodes among the 
$N$ evaluation episodes:
\begin{equation}
\overline{m}_{\text{succ}} =
\frac{\sum_{i=1}^N m_i \, \mathbf{1}[\text{succ}_i]}
     {\sum_{i=1}^N \mathbf{1}[\text{succ}_i]},
\qquad
m \in \{\tau, L_{\text{ee}}, L_{\text{joint}}, J, R\}.
\label{eq:average}
\end{equation}

\noindent Most efficient-VLA studies report performance in simulation, where environments are controllable and standardized. However, accurate modeling of actuator dynamics and heat dissipation remains a known limitation of simulators, making direct energy estimation unreliable~\cite{Hwangbo_2019}. Therefore, instead of simulated energy, we adopt the aforementioned set $(\overline{\tau}, \overline{L}_{\text{ee}}, \overline{L}_{\text{joint}}, \overline{J}, \overline{R})$ as kinematic and control-based proxies correlated with traversal distance and actuation effort, serving as informative indicators of energy efficiency. Taken together, $(\text{SR}, \overline{\tau}, \overline{L}_{\text{ee}}, \overline{L}_{\text{joint}}, \overline{J}, \overline{R})$ provide a task-aware, kinematics- and control-based perspective of efficiency: successful policies that complete tasks faster, with shorter paths and smoother motions, correspond to closed-loop behaviors that are more amenable to energy-efficient execution on physical robots.

\begin{table*}[t]
\caption{Evaluation of VLA models compressed with unstructured \ul{\textit{weight pruning}}~\cite{han2015deep} across embodied efficiency metrics averaged over four task suites of Libero (Libero-Spatial, Libero-Object, Libero-Goal, Libero-10). $\mathbf{\downarrow\!w}$ denotes the pruning ratio.
Norm. $\overline{\tau}$, $\overline{L}_{\text{ee}}$, $\overline{L}_{\text{joint}}$, $\overline{J}$, and $\overline{R}$ indicate the normalized task completion time, end-effector path length, joint-space path length, average jerk L2 norm, and average action rate relative to the baseline, respectively.}
\vspace{-0.7em}
\centering
\resizebox{\linewidth}{!}{
\begin{tabular}{l||c|c|c||c|c|c||c|c|c|c}
\toprule
\multirow{2}{*}{\textbf{Model}} & \multicolumn{3}{c||}{\textbf{$\mathbf{\pi_0}$}} & \multicolumn{3}{c||}{\textbf{$\mathbf{\pi_{0.5}}$}} & \multicolumn{4}{c}{\textbf{MolmoAct}} \\
& Baseline & $\downarrow\!w = 5\%$& $\downarrow\!w = 10\%$ & Baseline & $\downarrow\!w = 5\%$ & $\downarrow\!w = 10\%$ & Baseline & $\downarrow\!w = 5\%$ & $\downarrow\!w = 10\%$ & $\downarrow\!w = 20\%$ \\
\midrule
Success Rate (\%) & 91.3 & 92.1 (+0.8) & 88.6 (-2.7) & 96.8 & 96.7 (+0.1) & 96.1 (-0.7) & 86.3 & 86.5 (+0.2) & 88.8 (+2.5) & 87.3 (+1.0) \\
\midrule
Norm. $\overline{\tau}$ (\%) & 100 & 99.9 (-0.1) & 101.6 (+1.6) & 100 & 100.8 (+0.8) & 103.0 (+3.0) & 100 & 99.0 (-1.0) & 101.2 (+1.2) & 101.5 (+1.5) \\
\midrule
Norm. $\overline{L}_{\text{ee}}$ (\%) & 100 & 101.7 (+1.7) & 105.6 (+5.6) & 100 & 100.2 (+0.2) & 100.0 (+0.0) & 100 & 101.5 (+1.5) & 100.0 (+0.0) & 102.3 (+2.3) \\
\midrule
Norm. $\overline{L}_{\text{joint}}$ (\%) & 100 & 99.0 (-1.0) & 102.8 (+2.8) & 100 & 100.4 (+0.4) & 100.0 (+0.0) & 100 & 100.4 (+0.4) & 101.4 (+1.4) & 100.7 (+0.7) \\
\midrule
Norm. $\overline{J}$ (\%) & 100 & 102.6 (+2.6) & 111.0 (+11.0) & 100 & 101.6 (+1.6) & 100.7 (+0.7) & 100 & 96.8 (-3.2) & 101.9 (+1.9) & 105.8 (+5.8) \\
\midrule
Norm. $\overline{R}$ (\%) & 100 & 100.8 (+0.8) & 103.1 (+3.1) & 100 & 98.0 (-2.0) & 102.5 (+2.5) & 100 & 99.2 (-0.8) & 100.1 (+0.1) & 101.0 (+1.0) \\
\bottomrule
\end{tabular}
}
\label{tab:prune}
\end{table*}

\section{Inference Efficiency vs. Embodied Efficiency}
\label{sec:exp:embodied}

With the embodied efficiency metrics defined in Sec.~\ref{sec:preliminaries:metrics}, we conduct experiments on compressing VLA models across different domains, i.e., model/token/action, to investigate how compression techniques originally designed to improve inference efficiency affect embodied efficiency. For each task in the benchmark suite, i.e., Libero-Spatial, Libero-Object, Libero-Goal, Libero-10~\cite{liu2023libero}, and Bridge~\cite{walke2024bridgedatav2datasetrobot}, we run $N = 50$ evaluation episodes, consistent with the settings in OpenPi~\cite{openpi2025}, or until 10 successful episodes ($\mathbf{1}[\text{succ}_i] = 1$ for the $i$-th) are obtained, following the settings in the original Libero dataset~\cite{liu2023libero}. We report metrics averaged across episodes using Eq.~\ref{eq:average}.

\subsection{Experiments on Model Compression}
\label{sec:exp:embodied:model}

For model compression, we conduct experiments using weight pruning~\cite{han2015deep} and quantization~\cite{frantar2022gptq} on several representative VLA models, including $\pi_{0}$~\cite{black2024pi0}, $\pi_{0.5}$~\cite{black2025pi05}, and MolmoAct~\cite{lee2025molmoact}. 

\begin{figure}[t] 
    \centering \includegraphics[width=1\linewidth]{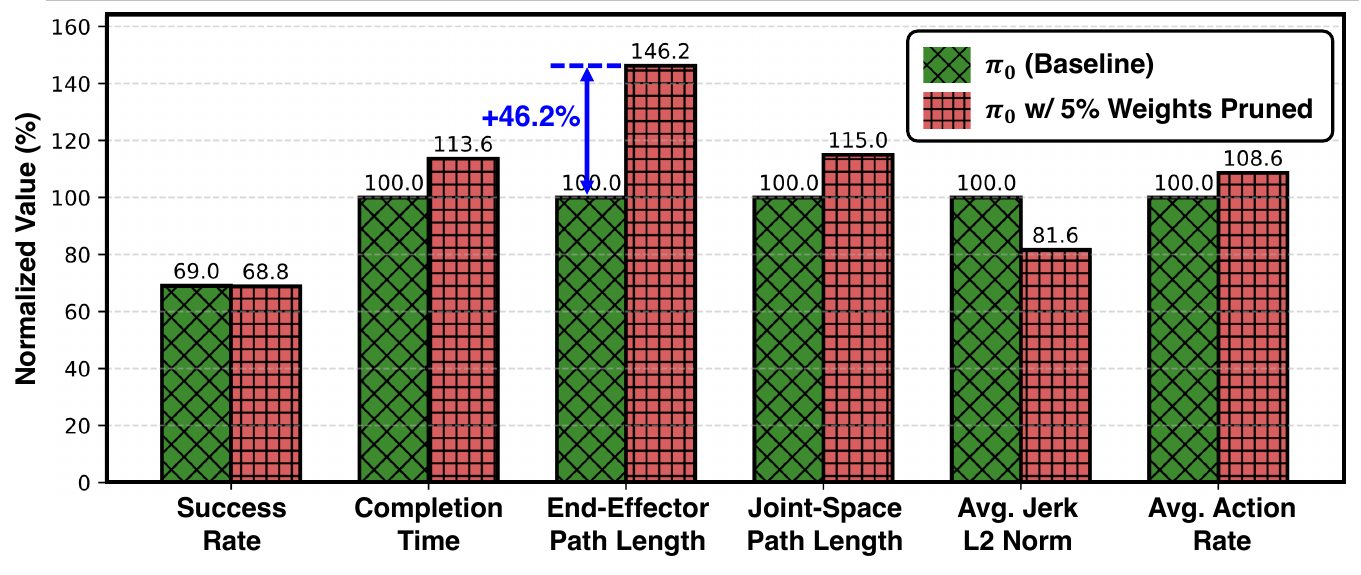} 
        \vspace{-2.5em}
\caption{Comparison of the baseline $\pi_{0}$~\cite{black2024pi0,fang2025intention} and its pruned variant (5\% weights removed via magnitude pruning~\cite{han2015learning}) on the Bridge benchmark~\cite{walke2024bridgedatav2datasetrobot}.}
\vspace{-1.2em}
    \label{fig:bridge} 
\end{figure}

\noindent \textbf{Weight pruning.} As summarized in Tab.~\ref{tab:prune}, we prune the weights with the smallest magnitudes in each baseline model, where $\mathbf{\downarrow\!w}$ denotes the pruning rate. We observe that when the pruning rate $\mathbf{\downarrow\!w}$ ranges from 5\% to 20\%, without notably hurting the task success rate (e.g., at most -2.7\% drop), the end-to-end execution cost increases noticeably (e.g., up to +5.6\% in terms of~\text{end-effector path length}), and motion smoothness degrades notably (e.g., up to +11.0\% in average jerk L2 norm). To understand the decreased embodied efficiency, we further visualize the trajectories of a baseline model and a pruned model in Fig.~\ref{fig:example_traj}. Although both models complete the task successfully, the pruned model follows a noticeably longer path, an aspect that is difficult to capture with conventional model inference efficiency metrics. These results indicate that although pruning effectively reduces model size, it can make the model less \emph{embodied-efficient} than uncompressed baselines. We further conduct unstructured weight pruning on a more challenging benchmark, Bridge~\cite{walke2024bridgedatav2datasetrobot}. As shown in Fig.~\ref{fig:bridge}, even with only 5\% of the weights pruned, the end-effector path length increases by 46.2\% with only a 0.2\% drop in success rate, consistent with our observations in Tab.~\ref{tab:prune}.

\textbf{Weight quantization.} As summarized in Tab.~\ref{tab:quant}, we explore quantizing the \texttt{bfloat16} models to \texttt{int8} and \texttt{int4}. Similar to our observations on model pruning, while quantization largely preserves task success rate, with the worst-case drop limited to $-1.8\%$, 
it introduces degradation in motion smoothness. The average jerk L2 norm $\bar{J}$ increases across all VLA models under both quantization levels, with \texttt{int4} quantization yielding particularly pronounced jerk increases of up to $+19.5\%$ ($\pi_0$), $+14.1\%$ (MolmoAct), and $+3.5\%$ ($\pi_{0.5}$).

\begin{table*}[!t]
\caption{
Evaluation of VLA models compressed with post-training \ul{\textit{weight quantization}}~\cite{frantar2022gptq} across embodied efficiency metrics averaged over four task suites of Libero (Libero-Spatial, Libero-Object, Libero-Goal, Libero-10). $\mathbf{P(w)}$ denotes the weight bit precision.
Norm. $\overline{\tau}$, $\overline{L}_{\text{ee}}$, $\overline{L}_{\text{joint}}$, $\overline{J}$, and $\overline{R}$ indicate the normalized task completion time, end-effector path length, joint-space path length, average jerk L2 norm, and average action rate relative to the baseline, respectively. 
Following the official OpenPi implementation \cite{openpi2025}, we employ a fake quantization scheme to evaluate the impact of bit-precision reduction. Specifically, weights are first quantized to the target precision $\mathbf{P(w)}$ and subsequently dequantized back to their original floating-point format to simulate the quantization error during inference, ensuring compatibility with the model's standard execution pipeline.}

\vspace{-0.7em}
\centering
  \resizebox{\linewidth}{!}
  {
    \begin{tabular}{l||c|c|c||c|c|c||c|c|c}
    \toprule
    \multirow{2}{*}{\textbf{Metric}} & \multicolumn{3}{c||}{\textbf{$\mathbf{\pi_0}$}} & \multicolumn{3}{c||}{\textbf{$\mathbf{\pi_{0.5}}$}} & \multicolumn{3}{c}{\textbf{MolmoAct}} \\
    & Baseline & $\mathbf{P(w)} = \texttt{int8}$ & $\mathbf{P(w)} = \texttt{int4}$ & Baseline & $\mathbf{P(w)} = \texttt{int8}$ & $\mathbf{P(w)} = \texttt{int4}$ & Baseline & $\mathbf{P(w)} = \texttt{int8}$ & $\mathbf{P(w)} = \texttt{int4}$ \\
\midrule
Success Rate (\%) & 92 & 92.7 (+0.7) & 90.2 (-1.8) & 96.3 & 96.3 (+0.0) & 97.1 (+0.8) & 86.3 & 88.2 (+1.9) & 88.6 (+2.3) \\
\midrule
Norm. $\overline{\tau}$ (\%) & 100 & 99.8 (-0.2) & 102.8 (+2.8) & 100 & 100.3 (+0.3) & 99.3 (-0.7) & 100 & 100.9 (+0.9) & 100.7 (+0.7) \\
\midrule
Norm. $\overline{L}_{\text{ee}}$ (\%) & 100 & 99.9 (-0.1) & 101.1 (+1.1) & 100 & 100.0 (+0.0) & 99.6 (-0.4) & 100 & 100.0 (+0.0) & 100.0 (+0.0) \\
\midrule
Norm. $\overline{L}_{\text{joint}}$ (\%) & 100 & 100.6 (+0.6) & 104.4 (+4.4) & 100 & 100.1 (+0.1) & 99.2 (-0.8) & 100 & 103.1 (+3.1) & 101.1 (+1.1) \\
\midrule
Norm. $\overline{J}$ (\%) & 100 & 100.4 (+0.4) & 119.5 (+19.5) & 100 & 96.6 (-3.4) & 103.5 (+3.5) & 100 & 100.8 (+0.8) & 114.1 (+14.1) \\
\midrule
Norm. $\overline{R}$ (\%) & 100 & 100.2 (+0.2) & 104.1 (+4.1) & 100 & 99.6 (-0.4) & 100.1 (+0.1) & 100 & 100.2 (+0.2) & 101.8 (+1.8) \\
\bottomrule
    \end{tabular}
    }
  \label{tab:quant}
\end{table*}

Overall, these results suggest that even when VLA models with compressed weights achieve task success rates comparable to their uncompressed counterparts, their \textbf{embodied efficiency} can differ substantially due to variations in motion dynamics, trajectory smoothness, and overall actuation effort. These system-level factors, which directly affect execution time and energy usage, are not reflected in conventional \textbf{inference efficiency} metrics that focus solely on computational cost during model forward passes.

\begin{figure*}[!t] 
    \centering \includegraphics[width=1\linewidth]{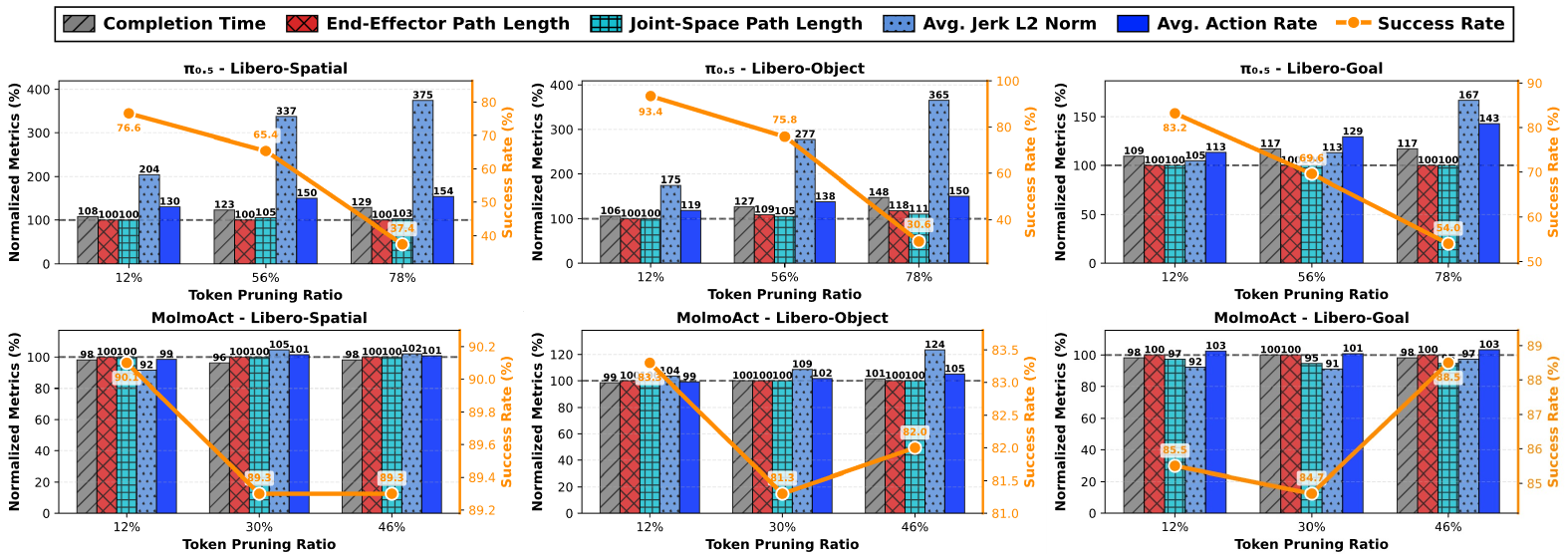 } 
        \vspace{-2em}
\caption{Evaluation of VLA models compressed with visual \ul{\textit{token pruning}}~\cite{yang_efficientvla_2025} across different embodied efficiency metrics. Each subplot corresponds to a specific \emph{model-task-suite} pair (e.g., ``$\pi_{0.5}$ - Libero-Goal'' represents evaluating the $\pi_{0.5}$ model~\cite{black2025pi05} on the Libero-Goal~\cite{liu2023libero} task suite under different token pruning ratios). All embodied efficiency metrics are shown as percentages normalized to each model’s unpruned baseline.}
    \label{fig:token} 
\end{figure*}

\subsection{Experiments on Token Compression}
\label{sec:exp:embodied:token}

Besides directly compressing model weights via pruning or quantization in Sec.~\ref{sec:exp:embodied:model}, we also explore applying the visual token pruning technique~\cite{yang_efficientvla_2025} to $\pi_{0.5}$~\cite{black2025pi05} and MolmoAct~\cite{lee2025molmoact}. As depicted in Fig.~\ref{fig:token}, we observe that higher token pruning ratios often lead to jerkier trajectories. For example, when the pruning ratio increases from 12\% to 56\% and 78\%, the average jerk L2 norm rises from 204\% to 337\% and 375\% of the unpruned model for the $\pi_{0.5}$~\cite{black2025pi05} model evaluated on the Libero-Spatial~\cite{liu2023libero} task suite, highlighting the negative impact of token-level compression on embodied efficiency.

\subsection{Experiments on Action Compression}
\label{sec:exp:embodied:action}
In addition to model- and token-level compression examined in Sec.~\ref{sec:exp:embodied:model} and Sec.~\ref{sec:exp:embodied:token}, respectively, we further investigate compression in the \emph{action space}. Following the settings in~\cite{pertsch2025fastefficientactiontokenization}, we replace the action head in $\pi_{0}$~\cite{black2024pi0} with the FAST action tokenizer~\cite{pertsch2025fastefficientactiontokenization}. As shown in Tab.~\ref{tab:action}, FAST generates less smooth actions, even increasing the average jerk L2 norm by 28.0\% to 50.6\%, while reducing completion time by 1.5\% to 5.6\%. These trade-offs, revealed by embodied efficiency metrics, offer a more comprehensive understanding of how different action tokenizers balance execution speed, motion smoothness, and energy efficiency. Such insights can guide practitioners to select action tokenization strategies that best align with the requirements and constraints of their specific robotic applications.

\begin{table*}[!t]
\caption{Evaluation of VLA models compressed with temporal \ul{\textit{action compression}} via the FAST tokenizer~\cite{pertsch2025fastefficientactiontokenization} across different embodied efficiency metrics.}
\vspace{-0.7em}
\centering
  \resizebox{\linewidth}{!}
  {\setlength{\tabcolsep}{12pt}
    \begin{tabular}{c|c||c||c|c|c||c|c}
    \toprule
    \multirow{2}{*}{\textbf{Task Suite}} & \multirow{2}{*}{\textbf{Tokenizer}}  & \multirow{2}{*}{\textbf{Success Rate}} & \textbf{Completion} & \textbf{End-Effector} & \textbf{Joint-Space} & \textbf{Avg. Jerk } & \textbf{Avg. Action} \\
    &   &  &  \textbf{Time $\overline{\tau}$} & \textbf{Path Length $\overline{L}_{\text{ee}}$} & \textbf{Path Length $\overline{L}_{\text{joint}}$} & \textbf{L2 Norm $\overline{J}$} & \textbf{Rate $\overline{R}$} \\
    \midrule
    \multirow{2}{*}{Libero-Spatial} 
    & $\pi_{0}$~\cite{black2024pi0} & 
    99.1\% & 5.4 & 1.0 & 4.0 & 861.9 & 0.124 \\
    & FAST~\cite{pertsch2025fastefficientactiontokenization} & 
    96.0\% (-3.1) & 5.1 (-5.6\%) & 1.0 (+0.0\%) & 3.8 (-5.0\%) & 1298.2 (+50.6\%) & 0.125 (+0.8\%) \\
    \midrule
    \multirow{2}{*}{Libero-Object} 
    & $\pi_{0}$~\cite{black2024pi0} & 
    99.1\% & 6.8 & 1.0 & 3.8 & 464.5 & 0.106 \\
    & FAST~\cite{pertsch2025fastefficientactiontokenization} & 
    99.0\% (-0.1) & 6.7 (-1.5\%) & 1.1 (+10.0\%) & 3.7 (-2.6\%) & 624.6 (+34.5\%) & 0.104 (-1.9\%) \\
    \midrule
    \multirow{2}{*}{Libero-Goal} 
    & $\pi_{0}$~\cite{black2024pi0} & 
    95.0\% & 5.6 & 0.9 & 3.6 & 1540.9 & 0.114 \\
    & FAST~\cite{pertsch2025fastefficientactiontokenization} & 
    91.2\% (-3.8) & 5.4 (-3.6\%) & 0.8 (-11.1\%) & 3.6 (+0.0\%) & 1973.1 (+28.0\%) & 0.112 (-1.8\%) \\
    \bottomrule
    \end{tabular}
    \vspace{1em}
    }
  \label{tab:action}
\end{table*}

\begin{table*}[!t]
\caption{Evaluating the $\pi_{0.5}$ VLA model~\cite{black2025pi05} under different \ul{\textit{in-context learning}} prompt designs. This implementation adopts the official $\pi_{0.5}$ configuration~\cite{openpi2025} to support the use of in-context prompts.
}
\vspace{-0.7em}
\centering
  \resizebox{\linewidth}{!}
  {\setlength{\tabcolsep}{11pt}
    \begin{tabular}{c|c||c||c|c|c||c|c}
    \toprule
     \multirow{2}{*}{\textbf{Task Suite}} & \multirow{2}{*}{\textbf{Prompt Type}}  & \multirow{2}{*}{\textbf{Success Rate}} & \textbf{Completion} & \textbf{End-Effector} & \textbf{Joint-Space} & \textbf{Avg. Jerk } & \textbf{Avg. Action} \\
    &   &  &  \textbf{Time $\overline{\tau}$} & \textbf{Path Length $\overline{L}_{\text{ee}}$} & \textbf{Path Length $\overline{L}_{\text{joint}}$} & \textbf{L2 Norm $\overline{J}$} & \textbf{Rate $\overline{R}$} \\
    \midrule
    \multirow{3}{*}{Libero-Spatial} & None & 98.4\% & 5.2 & 1.0 & 3.9 & 678.4 & 0.122 \\
    & Brief & 98.4\% (+0.0) & 5.4 (+2.3\%) & 1.0 (+0.0\%) & 3.8 (-1.2\%) & 628.1 (-7.4\%) & 0.119 (-2.5\%) \\
    & Detail & 98.2\% (-0.2) & 5.8 (+11.2\%) & 1.0 (+0.0\%) & 3.9 (+0.0\%) & 587.5 (-13.4\%) & 0.112 (-8.2\%) \\
    \midrule
    \multirow{3}{*}{Libero-Object} & None & 97.6\% & 7.0 & 1.1 & 3.9 & 506.8 & 0.118 \\
    & Brief & 98.8\% (+1.2) & 7.0 (+0.0\%) & 1.1 (+0.0\%) & 3.8 (-2.0\%) & 462.2 (-8.8\%) & 0.115 (-2.54\%) \\
    & Detail & 98.0\% (+0.4) & 7.3 (+4.4\%) & 1.1 (+0.0\%) & 3.7 (-4.1\%) & 409.3 (-19.24\%) & 0.112 (-5.1\%) \\
    \midrule
    \multirow{3}{*}{Libero-Goal} & None & 97.2\% & 5.6 & 0.8 & 3.7 & 1524.9 & 0.116 \\
    & Brief & 96.8\% (-0.4) & 5.6 (+0.0\%) & 0.8 (+0.0\%) & 3.7 (+0.0\%) & 1290.3 (-15.4\%) & 0.110 (-5.1\%) \\
    & Detail  & 96.2\% (-1.0) & 5.8 (+4.3\%) & 0.8 (+0.0\%) & 3.6 (-2.0\%) & 1131.9 (-25.8\%) & 0.105 (-9.5\%) \\
     \bottomrule
    \end{tabular}
    \vspace{1em}
    }
  \label{tab:new_in_context}
\end{table*}

\begin{table*}[!t]
\caption{Evaluating the $\pi_{0.5}$ VLA model~\cite{black2025pi05} under different \ul{\textit{supervised fine-tuning}} loss designs on the Libero~\cite{liu2023libero} dataset.}
\vspace{-0.7em}
\centering
  \resizebox{\linewidth}{!}
  {\setlength{\tabcolsep}{11pt}
    \begin{tabular}{c|c||c||c|c|c||c|c}
    \toprule
    \multirow{2}{*}{\textbf{Task Suite}} & \multirow{2}{*}{\textbf{Training Loss}}  & \multirow{2}{*}{\textbf{Success Rate}} & \textbf{Completion} & \textbf{End-Effector} & \textbf{Joint-Space} & \textbf{Avg. Jerk } & \textbf{Avg. Action} \\
    &   &  &  \textbf{Time $\overline{\tau}$} & \textbf{Path Length $\overline{L}_{\text{ee}}$} & \textbf{Path Length $\overline{L}_{\text{joint}}$} & \textbf{L2 Norm $\overline{J}$} & \textbf{Rate $\overline{R}$} \\
    \midrule
    \multirow{2}{*}{Libero-Spatial} 
    & Baseline & 100.0\% & 5.2 & 1.0 & 3.9 & 631.3 & 0.122 \\
    & + Auxiliary Loss & 99.6\% (-0.4) & 5.2 (+0.0\%) & 1.0 (+0.0\%) & 3.8 (-2.6\%) & 526.6 (-16.6\%) & 0.115 (-5.7\%) \\
    \midrule
    \multirow{2}{*}{Libero-Object} 
    & Baseline & 99.1\% & 6.7 & 1.1 & 3.8 & 404.7 & 0.107 \\
    & + Auxiliary Loss & 99.4\% (+0.3) & 6.9 (+3.0\%) & 1.1 (+0.0\%) & 3.8 (+0.0\%) & 351.3 (-13.2\%) & 0.100 (-6.5\%) \\
    \midrule
    \multirow{2}{*}{Libero-Goal} 
    & Baseline & 96.4\% & 5.3 & 0.8 & 3.6 & 1810.0 & 0.112 \\
    & + Auxiliary Loss & 97.6\% (+1.2) & 5.5 (+3.8\%) & 0.8 (+0.0\%) & 3.6 (+0.0\%) & 1374.9 (-24.0\%) & 0.104 (-7.1\%) \\
    \midrule
    \end{tabular}
    }
  \label{tab:sft}
\end{table*}

\section{Evaluating Adaptation Strategies in Terms of Embodied Efficiency}
\label{sec:exp:adapt}

Motivated by the unique perspective offered by embodied efficiency metrics (see Sec.~\ref{sec:exp:embodied}) compared to conventional evaluations, we further investigate whether standard adaptation strategies used in large language models can effectively steer VLA behavior toward higher embodied efficiency. We explore two approaches. (1) In-context learning~\cite{dong2024surveyincontextlearning}: we augment the prompt with optimization targets based on the embodied-efficiency metrics defined in Sec.~\ref{sec:preliminaries:metrics}, enabling the model to condition its predictions on efficiency-aware context. (2) Supervised fine-tuning: we incorporate an embodied-efficiency term as an auxiliary loss to regularize action generation.

\subsection{Experiments on In-Context Learning}
We prompt the $\pi_{0.5}$ model using two forms of efficiency-aware context, \textit{Brief} and 
\textit{Detail}, and evaluate the resulting action quality across three task suites. Notably, we conduct these experiments on $\pi_{0.5}$ (rather than $\pi_0$) because the original $\pi_0$ model has a constrained maximum language token length that precludes the use of long-form in-context prompts. In the 
\textit{Brief} setting, the original task prompt is augmented with a high-level directive: 
\textit{``Generate the actions to minimize energy consumption.''} In the \textit{Detail} setting, 
we provide a more explicit optimization objective: \textit{``To minimize energy consumption, minimize 
the action rate $R = \frac{1}{T-1} \sum_{t=1}^{T-1} \lVert a_{t+1} - a_t \rVert_2$.''} We also instruct 
the model to reduce the total number of steps. As shown in Tab.~\ref{tab:new_in_context}, in-context learning consistently reduces the average action rate $\bar{R}$ across all three task suites under both prompt designs, with substantial reductions in the average jerk L2 norm $\bar{J}$ across all task suites. These reductions, however, come at the cost of longer completion time, with increases of up to $+11.2\%$.

\subsection{Experiments on Supervised Fine-Tuning}
We fine-tune the $\pi_{0.5}$ VLA model~\cite{black2025pi05} using an auxiliary Jerk-L2 loss and Action-Rate loss; that is, we add $\eta \overline{J}$ and $\eta \overline{R}$ ($\eta = 0.01$) during training to balance the magnitudes of the auxiliary loss and the baseline loss. As shown in Tab.~\ref{tab:sft}, this loss design reduces the average jerk across different task suites by 16.6\% to 24.0\% without compromising success rate (+0.4\% on average). However, the corresponding task completion time increases by up to 3.8\%.

\section{Conclusion}
In this work, we show that conventional inference-centric metrics do not reliably predict real-world execution quality for VLA models. Methods that appear efficient under standard evaluations can lengthen task duration, increase path length, and degrade motion smoothness. Dedicated embodied efficiency metrics can expose differences that remain hidden under conventional evaluations alone. Additionally, common adaptation methods provide only mild and metric-specific improvements in embodied efficiency, often with trade-offs. Our results highlight the need for efficient VLA research to jointly consider inference efficiency and embodied execution. Incorporating the investigated embodied-efficiency metrics in this work has the potential to enable a more faithful assessment of closed-loop behavior, guiding the development of VLA systems that are both computationally efficient and behaviorally robust.

{
    \small
    \bibliographystyle{IEEEtran}
    \bibliography{main}
}

\end{document}